\documentclass{article}

\PassOptionsToPackage{numbers, compress}{natbib}

\usepackage[preprint]{nips_2018}

\usepackage[utf8]{inputenc} 
\usepackage[T1]{fontenc}    
\usepackage{hyperref}       
\usepackage{url}            
\usepackage{booktabs}       
\usepackage{amsfonts}       
\usepackage{nicefrac}       
\usepackage{microtype}      
\usepackage{wrapfig}

\usepackage{color}
\usepackage{listings}
\usepackage{gensymb}
\usepackage{subcaption}
\usepackage{graphicx}
\usepackage{enumitem}

\newcommand{\ppo}{\texttt{PPO2}}

\definecolor{mygreen}{rgb}{0,0.6,0}
\definecolor{mygray}{rgb}{0.5,0.5,0.5}
\definecolor{mymauve}{rgb}{0.58,0,0.82}
\definecolor{myred}{rgb}{0.6,0.1,0.1}
\newcommand{\toybox}{\textsc{ToyBox}}

\title{\toybox{}: Better Atari Environments for \\Testing  Reinforcement Learning Agents}

\newcommand{\kaleigh}{    
  Kaleigh Clary \\
  College of Information and Computer Science\\
  University of Massachusetts Amherst\\
  Amherst, MA 01060 \\
  \texttt{kclary@cs.umass.edu}
}

\newcommand{\john}{
  John Foley* \\
  Department of Computer Science \\
  Smith College \\
  Northampton, MA 01060 \\
  \texttt{jjfoley@smith.edu} 
}

\newcommand{\david}{
    David Jensen \\
  College of Information and Computer Science\\
  University of Massachusetts Amherst\\
  Amherst, MA 01060 \\
    \texttt{jensen@cs.umass.edu}    
}

\newcommand{\emma}{
Emma Tosch\thanks{Authors contributed equally.}\\
  College of Information and Computer Science\\
  University of Massachusetts Amherst\\
  Amherst, MA 01060 \\
\texttt{etosch@cs.umass.edu}
}

\author{
\john{}\\
\And
\emma{}
\And
\kaleigh{}\\
  \And
\david{}\\
}

\begin{document}

\maketitle

\begin{abstract}
It is a widely accepted principle that software without tests has bugs. Testing reinforcement learning agents is especially difficult because of the stochastic nature of both agents and environments, the complexity of state-of-the-art models, and the sequential nature of their predictions. Recently, the Arcade Learning Environment (ALE) has become one of the most widely used benchmark suites for deep learning research, and state-of-the-art Reinforcement Learning (RL) agents have been shown to routinely equal or exceed human performance on many ALE tasks. Since ALE is based on \textit{emulation} of original Atari games, the environment does not provide semantically meaningful representations of internal game state. This means that ALE has limited utility as an environment for supporting testing or model introspection. We propose \toybox{}, a collection of reimplementations of these games that solves this critical problem and enables robust testing of RL agents.
\end{abstract}

\section{Introduction}
\label{sec:intro}

Advances in deep models have fueled impressive gains in Reinforcement Learning (RL). Agents trained using deep RL can learn control policies for a variety of complex tasks using only sensory input~\cite{arulkumaran2017brief,hwangbo2017control}. One of the most impressive benchmarks is the set of Atari games served by the Arcade Learning Environment (ALE)~\cite{bellemare13arcade,machado17arcade,mnih2013playing} where deep models can often outperform human players. Consequently, Atari games have become the \textit{de facto} academic and industrial benchmarks for RL. Frameworks such as OpenAI Gym have further lowered the barrier to entry for training RL agents, and have made popular models available off the shelf~\cite{1606.01540,baselines}. 

However, these original Atari emulations are functionally black boxes, making it highly impractical to view and manipulate internal game state for testing purposes. Since RL agents are learned software, it is imperative that we can test these black-box software agents. This is difficult when the environment is also both complex and opaque.

ALE was originally designed strictly as a benchmark suite to enable comparison of the generalization properties of RL models~\cite{bellemare13arcade}. In practice, however, it is now common for researchers to develop new RL models using these Atari games. In a field where there are few instances of non-trivial separate training and testing environments, this trend presents challenges for how to address ALE's original goal.

The central challenge for testing RL agent behavior is that typically both the agent and the environment are black boxes. When an RL agent achieves human or super-human performance on a complex task, researchers want to make claims about the reasoning power and utility of their trained agents. This requires the ability to form \emph{and test} behavioral hypotheses in the environment. Therefore, the ideal RL test environment is a white-box environment with easily manipulable and composable parameters that permit the expression of high-level concepts. Such an environment ought to avoid design biases that would make the environment more tractable for learning. 

Our core contribution is the development of \toybox{}, a suite of fully parameterizable, replicated Atari games. \toybox{} facilitates a host of analyses not previously possible in ALE, including, but not limited to:
\begin{itemize}[leftmargin=*]
    \item \textbf{Post-Training Acceptance Testing:} An agent that has learned a concept ought to be able to perform similarly in similar states. \toybox{} permits the programmatic generation of similar states. An agent can begin interacting with an intervened-upon \toybox{} implementation of a game at any point during gameplay. This means that we can accept or reject agents based on their ability to succeed in critical tasks under laboratory conditions.
    \item \textbf{Dynamic Analysis during Training:} RL training can take a considerable amount of wall-clock time. Therefore, the cost of getting stuck in a local minimum can be high~\cite{rlblogpost}. Dynamic analysis of agent behavior could help diagnose problematic reward functions or random seeding earlier rather than later.
    \item \textbf{Test Set Generation:} Because \toybox{} games are parameterizable, users can specify a family of games by varying some of the parameter values. The specific features of what that entails will depend on the task. These families of games would be specifically useful for curriculum learning.
\end{itemize}

In this paper, we focus on the first novel analysis made possible via \toybox{}: that of determining the degree to which a specific, learned agent meets performance expectations expressed in human-understandable terms.  To do this, we focus on \textit{Breakout}, a specific game implemented within both ALE and \toybox{}. Testing whether a learned agent meets these performance expectations would be virtually impossible in ALE, but such testing is quite simple in \toybox{}.


\paragraph{Example: Breakout.} With \toybox{}, we are able to convert speculation about RL agent behavior into testable \emph{behavioral requirements}. We focus on the game of Breakout, as some of the most high profile instances of anthropomorphization have been expressed about RL agents playing this game~\cite{mnih2015nature}. We identify and test three behavioral requirements related to the playing of Breakout by a black-box RL agent:

\begin{enumerate}[label=\textbf{R\arabic*}, leftmargin=*]
    \item\label{h:last_brick} An agent will be able to quickly eliminate each brick in isolation.
    \item\label{h:polar_starts} An agent will be able to hit the ball regardless of the angle of its initial trajectory. 
    \item\label{h:eztunnels} An agent will quickly open tunnels to the roof, eagerly seeking the high reward of the titular "breaking-out" behavior.
\end{enumerate}

These requirements are fairly straightforward, but it is not clear how to test them with ALE. Regardless of whether they correctly express what we mean when we say an agent has ``learned'' to play Breakout, \toybox{} makes it possible to express them and our core contribution is the ability to test that an agent meets requirements of this style. 

In order to conduct our experiments, we trained an RL agent using the recommended \ppo{} algorithm, and the default parameter settings available in the OpenAI Baselines implementation~\cite{schulman2017proximal,baselines}.
We find that evidence supports \ref{h:polar_starts}, but \ref{h:last_brick} is complicated by certain bricks being more challenging than others and \ref{h:eztunnels} is completely unsupported: our tested agents are only good at making one specific tunnel. Before \toybox{} these hypotheses were completely untestable in the existing ALE environment without significant time invested into reverse-engineering RAM locations.

\section{Background and Related Work}
Reinforcement learning is a subdomain of machine learning wherein an \emph{agent} interacts with an \emph{environment} via sequential \emph{actions}, seeking to maximize its expected cumulative \emph{reward}. The agent operates in the environment by taking actions to transition between environment \emph{states} ~\cite{sutton1998reinforcement}. 

Actions are typically selected according to a \emph{policy}, which is
a potentially stochastic function mapping states to actions. There are many methods for learning a policy, but essentially all RL agents learn by repeatedly acting in the environment and using the resulting experience to alter its policy. Deep reinforcement methods implement the policy function as a deep neural net. 

Arcade Learning Environment exports Atari states either as 160x210 pixel images, or as 1024 bit RAM. Actions in these environments are a set of inputs that can be applied to the joysticks, which have been discretized by ALE wrappers. The ALE designers have been able to identify the current game score in RAM, and extract it. Typical implementations of learners define the reward to be the truncated normalized score, i.e., +1, 0, -1 if the score has increased, not changed, or decreased, respectively. The agent is a computer program that returns joystick and button actions in response to Atari states.  

\paragraph{Simulation and Testing Environments.} 
Historically, RL researchers have created their own environments to test the behavior of RL algorithms. Environments drawn from the classic GridWorld family consist of an agent moving through a grid of states from a fixed start state toward a fixed goal state, typically with one or more obstacles or other complicating factors (e.g., teleporting states)~\cite{sutton1998reinforcement}. In this simplified world, the optimal policy is directly computable via dynamic programming.

Such simple environments are critically important for testing theoretical properties of RL agents' learned policies. Complex environments designed for human interaction do not typically have tractable ways of computing optimal policies. Among these more complicated environments, ALE is one of the most popular benchmarks, but real-time strategy games such as StarCraft, and simulations of real-world robotics problems using the MuJoCo physics engine, are also popular~\cite{arulkumaran2017brief, bellemare13arcade, synnaeve2016torchcraft, todorov2012mujoco, vinyals2017starcraft}.

Despite these recent achievements, the gap between toy problems and human domains remains quite large. Some authors have focused on generating families of theoretically understood games, which can help bridge this gap~\cite{raghu2017can}. Others have shown that, in sufficiently complex systems, the only feasible way to understand what an agent is learning is to build fast, intervenable systems~\cite{berseth2018terrain, lucas2018game, NIPS2017_6859}. 

The motivation for intervention has been largely about increasing variability during training. These systems all implemented either completely new games or simplified versions of existing games, having specialized parameterized implementations for the purposes of RL research. By re-implementing Atari games in \toybox{} we avoid any research-specific bias that might influence game design.

\paragraph{Challenges to Testing in Reinforcement Learning.}
In RL, it is often the case that the same environment is used for training, testing, and deployment. 
In fact, in the case of Atari games, testing and deployment are effectively the same: a researcher trains a new model using a particular game and then shows how well the agent learned by running the agent in that same environment. 

While there has certainly been criticism of the field for testing in the training environment~\cite{whiteson2011protecting}, there is a legitimate justification for doing so: if an agent is only to be deployed in the training environment, it is perfectly reasonable to overfit the agent's policy to that environment. However, claims about higher-level reasoning or generalization capabilities require testing, where testing means something more than simply running a trained agent in the environment.

Testing as construed in this paper is not the typical ``testing'' as referred to in supervised machine learning, where a classifier's ability to generalize is measured against a held-out evaluation set. Instead, we are arguing for a form of \textit{acceptance testing} where we take a set of trained RL agents and select the successful ones based on tests written to ensure conceptual accuracy. While we believe that this kind of testing and monitoring of models is present in industrial applications, there is little published work describing these procedures.

Testing in the software engineering sense is notoriously difficult for machine learning. The lack of determinism, combined with at times incredibly byzantine software layers and few theoretical guarantees, makes any kind of testing non-trivial. The closest analogue to unit testing would be training agents on a variety of classical problems, e.g., GridWorld experiments to show the capabilities of new algorithms. There are no common practices equivalent to integration, nor regression testing.


\section{System Design}

\toybox{} is a growing suite of Atari reimplementations, written in Rust. All games are written to be highly parameterized, with swappable components. Game state can be exported at any time, altered, and restarted from the intervened-upon state.

\begin{figure}
    \centering
    \includegraphics[width=0.49\columnwidth]{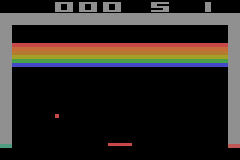}
    \includegraphics[width=0.49\columnwidth]{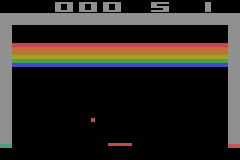}
    \caption{(\emph{Left}: ALE. \emph{Right}: \toybox{}.) Images of near-start frames for both Atari and \toybox{} implementations of Breakout. 
    }
    \label{fig:games}
\end{figure}

We have implemented OpenAI Gym environments that are drop-in replacements for the ALE OpenAI Gym implementations through \toybox{}'s Python API. This means that, for example, a user can train on the \toybox{} version of Breakout simply by registering and then loading the \toybox{} Breakout OpenAI environment. With care taken to ensure the same preprocessing steps are used on \toybox{} as on ALE Atari games, the same agents and networks can be trained on the testable \toybox{} interface instead.


Our fidelity to the original game has reached a level where the visual output between environments is nearly indistinguishable (see Figure~\ref{fig:games}), but remains a work in progress. We can already ask meaningful semantic questions through the \toybox{} API. In the future, we hope to improve our fidelity to the original games further, but since so much of Atari games is implemented in hardware and lookup tables, it is nearly impossible to code game physics that match Atari games bug-for-bug. While certain videogame bugs, such as the infamous overflow error in PacMan~\cite{pacmanOverflow}, are worth reproducing, we've focused this current work on testing desired properties on an environment that is simply indistinguishable by humans. 


\section{Semantic Tests}

As papers are published detailing new RL models for Atari playing, it is only natural to speculate as to the behaviors learned and understood by agents. \toybox{} provides us the ability to test these types of claims; here we test the three simple requirements (\ref{h:last_brick}, \ref{h:polar_starts}, \ref{h:eztunnels} from \S\ref{sec:intro}) as a demonstration of \toybox{}'s versatility and analytic power.

\paragraph{Shared Test Setup.}
All of the tests presented were run using two agents trained using the default settings for the \ppo{} algorithm available in OpenAI Gym Baselines~\cite{baselines, schulman2017proximal}. One agent was trained for 10 million steps (1e7) and the other for 50 million steps (5e7) in the \toybox{} Breakout environment. The 1e7 model achieves an average score of 407 and standard deviation of 86 points. The 5e7 model achieves an average score of 777, and a standard deviation of 154 points. In order to win Breakout, an agent must clear two identical levels, where each level is worth 432 points.   

We chose \ppo{} to use in our demonstration because this is one of OpenAI Gym's recommended algorithms\footnote{\url{https://blog.openai.com/openai-baselines-ppo/}} but our contributions in this paper are model-agnostic.


Many RL agents are stochastic in nature: during training, randomness is used to balance exploration and exploitation, while during deployment, a learned policy may need to break ties. This results in agents whose policies are stochastic. Therefore, for each experimental condition, we run 30 trials of each for the test conditions and present the aggregated results.

For each requirement, we ran many trials and constructed a software test that exited under conditions of both success and failure. Each trial was allocated the equivalent of four minutes of human gameplay time for the agent to reach a success or failure state. We gave the agent a single life in order to achieve its objective. Dying or failing to pass the test within the time span was considered a failure for all hypotheses. We allowed the agent only a single level of breakout, ending when all the bricks disappeared, rather than see if it can clear both levels for each test.

\paragraph{\ref{h:last_brick}: Brick Elimination.}
To win Breakout, an agent must eliminate every brick. A plausible state for such an agent might be after a ``death'' or lost ball, where there is one brick remaining. We might expect an agent that understands the physics of Breakout to remove the last brick quickly. 

\begin{figure}
  \begin{minipage}[b]{.33\linewidth}
    \centering\includegraphics[width=0.9\linewidth]{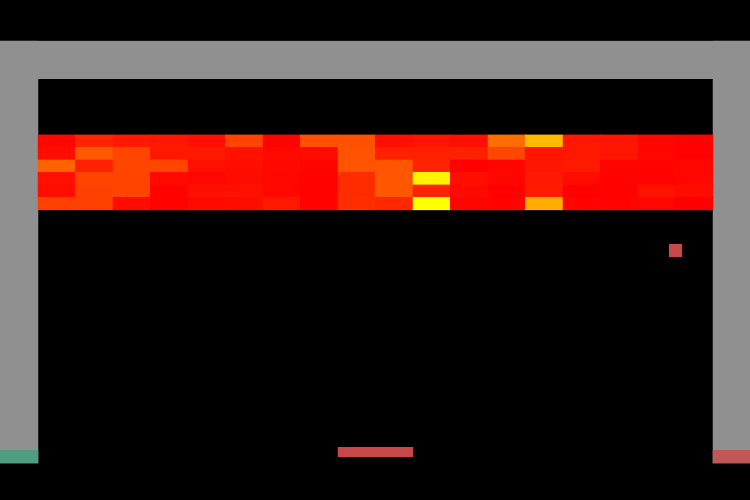}
    \subcaption{Brick Elimination (1e7)}\label{fig:last_brick_10}
  \end{minipage}%
  \begin{minipage}[b]{.33\linewidth}
    \centering\includegraphics[width=0.8\linewidth]{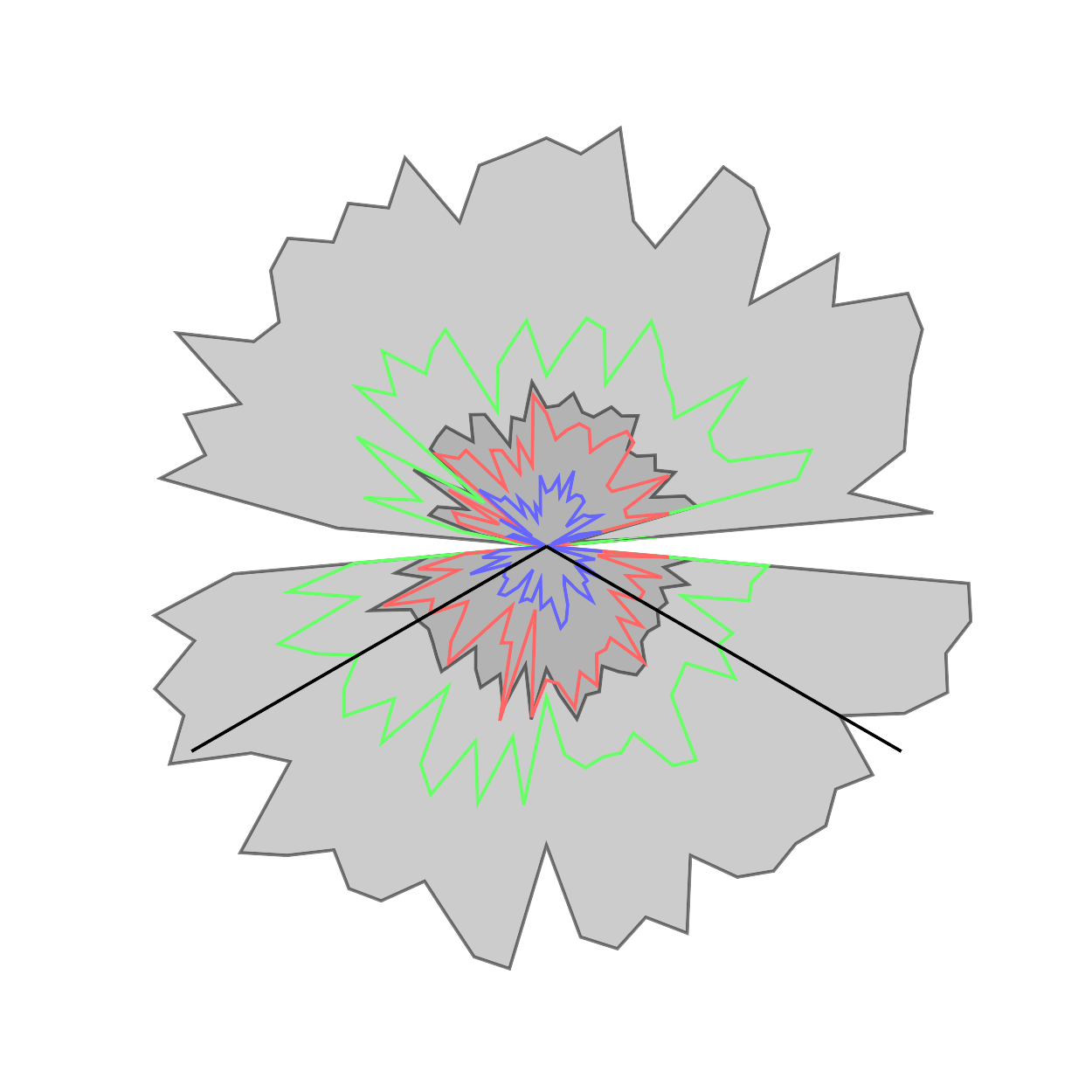}
    \subcaption{Start Angle Invariance (1e7)}\label{fig:polar_start_10}
  \end{minipage}
  \begin{minipage}[b]{.33\linewidth}
    \centering\includegraphics[width=0.9\linewidth]{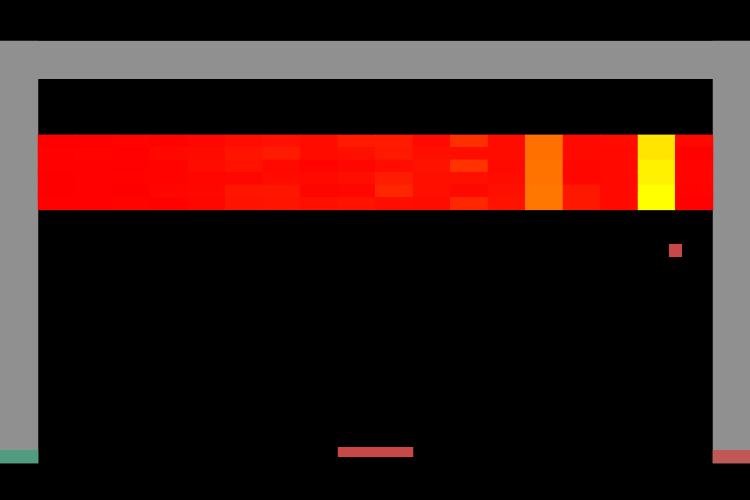}
    \subcaption{Tunnel Exploitation (1e7)}\label{fig:ez_tunnel_10}
  \end{minipage}
  \begin{minipage}[b]{.33\linewidth}
    \centering\includegraphics[width=0.9\linewidth]{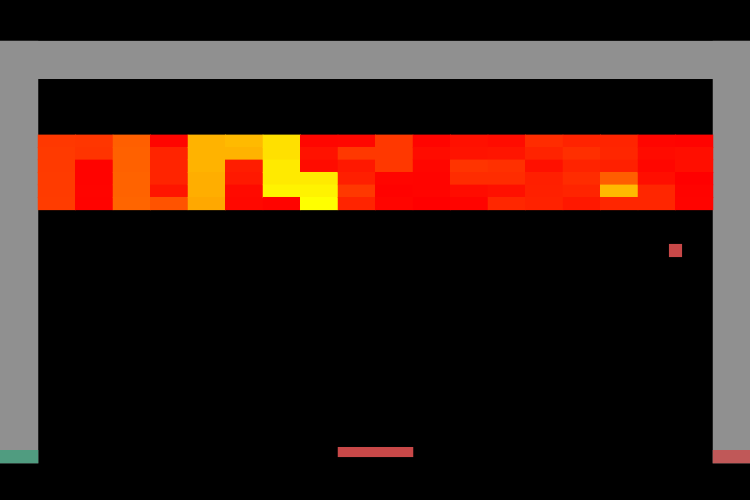}
    \subcaption{Brick Elimination (5e7)}\label{fig:last_brick_50}
  \end{minipage}%
  \begin{minipage}[b]{.33\linewidth}
    \centering\includegraphics[width=0.8\linewidth]{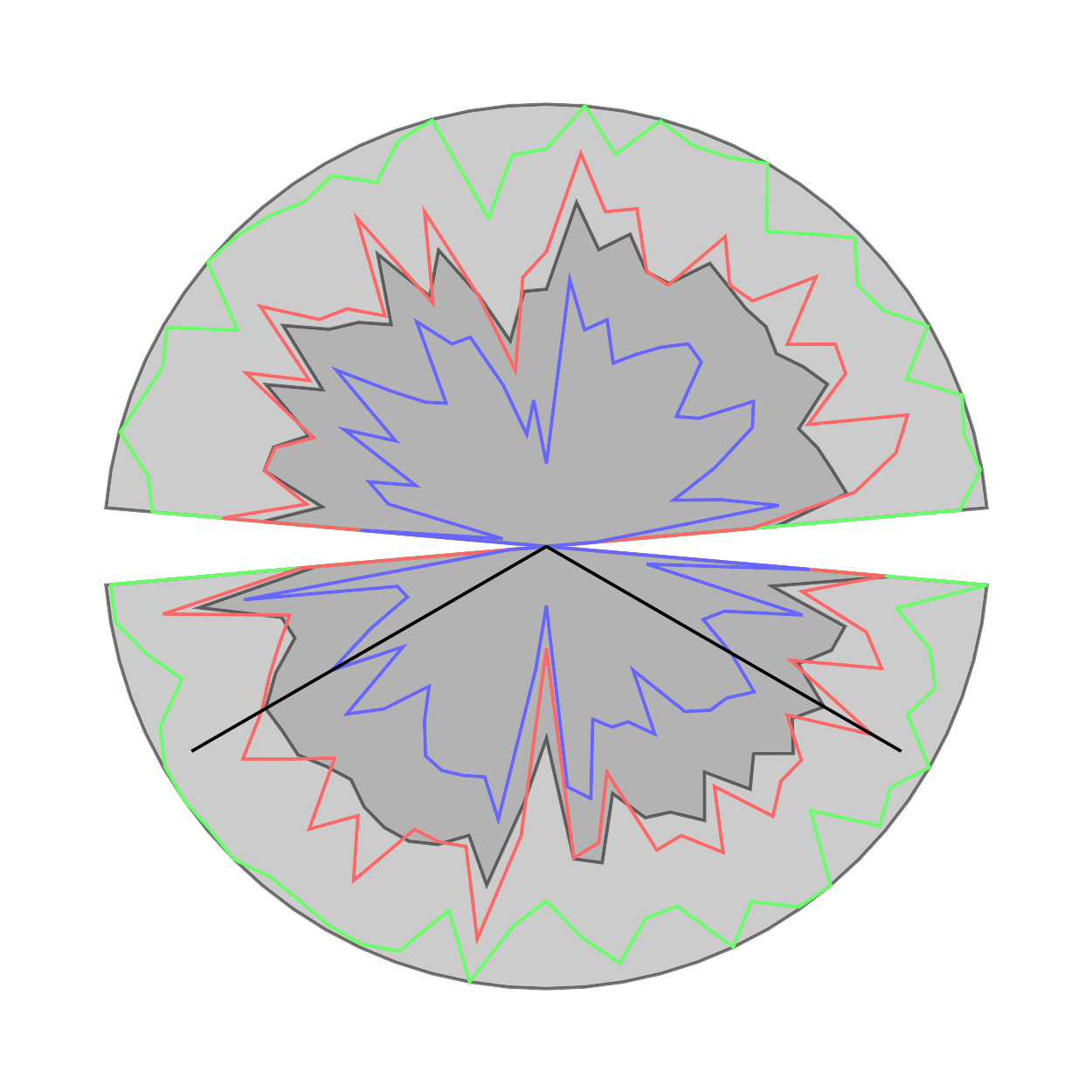}
    \subcaption{Start Angle Invariance (5e7)}\label{fig:polar_start_50}
  \end{minipage}
  \begin{minipage}[b]{.33\linewidth}
    \centering\includegraphics[width=0.9\linewidth]{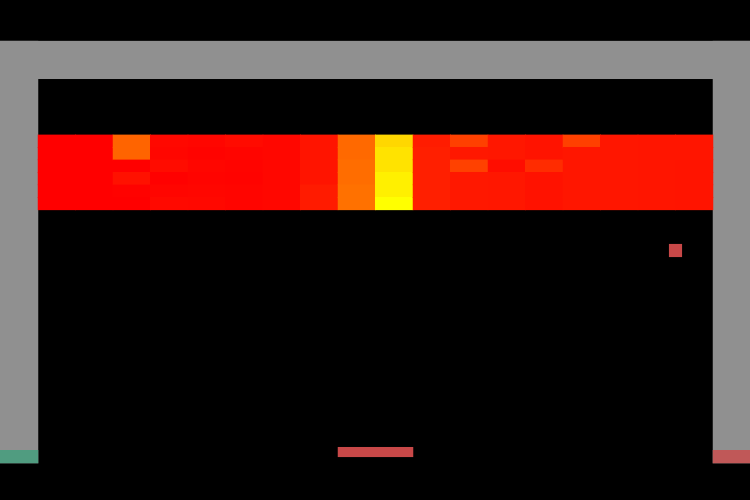}
    \subcaption{Tunnel Exploitation (5e7)}\label{fig:ez_tunnel_50}
  \end{minipage}
  \label{fig:viz}
  \caption{Test Data Visualizations: Color of bricks in Figures~\ref{fig:last_brick_10},  \ref{fig:ez_tunnel_10}, \ref{fig:last_brick_50},and \ref{fig:ez_tunnel_50} indicate the median number of steps required to clear the particular brick in that test: Bright yellow represents fewer steps and dark red indicate many more steps. In Figures~\ref{fig:polar_start_10} and \ref{fig:polar_start_50}, the black lines indicate the starting angles seen during training, the light gray area the maximum score achieved from this starting angle, and the dark gray area represents the mean score achieved across trials.}
\end{figure}

Figures~\ref{fig:last_brick_10} and \ref{fig:last_brick_50} depict the reciprocal of the median number of steps taken to eliminate a given brick, with yellow indicating very few steps (median of 20) and red indicating very many steps (median of 400). For example, the agent typically targeted the paddle so that the ball would hit straight up when it was launched from a start state.

We suspect that the large difference we see between brick elimination scenarios is that improvements in paddle aim are not evenly distributed spatially. The same temporal range is visible on both charts, (maximums and minimums have not shifted) but there are many more bricks that are eliminated quickly for the agent with more training steps.

We had expected to see, for example, symmetry across the bricks. However, there is no obvious pattern here, suggesting that the agents are not particularly good at aiming at final, lingering bricks. In the future, we hope to explore whether these latencies are related to the frequency of their occurrence as the last brick in training data. Some alternative hypotheses for explaining the agent's behavior are that it has memorized a sequence of actions, or that it has only learned to survive and is not targeting the ball at all. 

\paragraph{\ref{h:polar_starts}: Start Angle Invariance.}

Our next behavioral requirement comes from observation of the game. Over time, the ball bounces through many different angles, and it is plausible (though maybe not strictly possible in the original version based on fixed-point math) to see all possible angles of movement of the ball. We therefore expect an agent to be able to catch the ball ``thrown'' at any start angle and to resume play.

Since Breakout operates over a finite set of angles, we did not expect as much success as this test indicated. Modifying a start state to change the initial launch angle of the ball led to a single life of gameplay, and a total score. Over 30 trials per agent per angle, we are able to visualize the mean (dark gray area), max (light gray area), median (red), 25th (blue) and 75th (green) percentiles on a polar plot in Figures~\ref{fig:polar_start_10} and \ref{fig:polar_start_50}.

Both plots display failure to achieve any score with horizontal ball angles: since Breakout has no gravity, balls simply bounce horizontally forever, never hitting any bricks or threatening the paddle.


The agents under test were remarkably resilient to starting angles. While displaying a large amount of variance,\footnote{Although outside the scope of this work, there is a recent growing interest in the role of variability during evaluation in machine learning, as well as RL spcifically~\cite{cohen2018distributed, clary2018variability, jordan2018benchmarks}. We suspect the higher variance observed during evaluation of \ref{h:polar_starts} is an instance of this phenomenon.} the maximum score achieved from each angle was much higher, suggesting that an agent can be successful even with balls traveling at angles it may never have observed in training. The agent trained for 50 million steps was much more robust -- out of the 30 trials, there was at least one trial that completed the full level for each starting angle.

Looking closer at the mean, we can see that the agent trained for 5e7 steps had more difficulty with vertical angles. When we observed this behavior, the agent would sometimes keep the ball aligned perfectly in the center of the board, hitting it precisely in the center of the paddle, over and over, and therefore failing to make progress. This is a fascinating behavior that is entirely unlike the kind of behavior we would expect from human players.

\paragraph{\ref{h:eztunnels}: Tunnel Exploitation.} 
The highest reward in Breakout comes from ``digging'' a tunnel through the wall of bricks and then bouncing the ball through the hole and onto the ceiling. When the ball is trapped above the bricks, it will result in many bricks being removed very quickly, and a higher near-term reward than usual. Mnih et al., speculated that a high-performing agent was learning this behavior~\cite{mnih2015nature}.

One way to test whether an agent ``knows'' to exploit a tunnel is to give it a board with a nearly built tunnel, save for a single brick, and test whether the agent can aim at that single brick within a given amount of time. Just as in the test presented for \ref{h:last_brick}, we can generate a test for every brick. 

In Fig.~\ref{fig:ez_tunnel_10} and \ref{fig:ez_tunnel_50}, the value for each brick is the reciprocal of the median number of time steps before that brick was removed. The values range from 17 timesteps (bright-yellow) to 400 timesteps (dark red). 

In order to satisfy our requirement, we would expect that an agent could exploit a tunnel quickly anywhere on the board, however we find that agents hit the ball to predictable locations regardless of the board configuration. We feel comfortable rejecting the hypothesis behind our requirement. We can see as well that specific agents learn to hit the ball to specific locations -- the \ppo{} model trained for 10 million steps prefers to send the ball to the right side of the screen, and the agent trained for 50 million steps prefers to hit the ball to the center column.


\section{Conclusions}

We present \toybox{}, a reimplementation of popular Atari benchmarks that allow for acceptance testing, dynamic analysis during training, and test-set generation. We present three requirement studies over deep RL models trained on Breakout and make evidence-based claims about whether particular agents satisfy our high-level behavioral requirements.

\section*{Acknowledgements}
This material is based upon work supported by the United States Air Force under Contract No, FA8750-17-C-0120.  Any opinions, findings and conclusions or recommendations expressed in  this material are those of the author(s) and do not necessarily reflect the views of the United States Air Force.
\small
\bibliographystyle{plain}

\end{document}